\documentclass{article} % For LaTeX2e
\usepackage{iclr2018_workshop,times}
\usepackage{hyperref}
\usepackage{url}
\usepackage{graphicx}
\usepackage{booktabs} 
\usepackage{adjustbox}
\usepackage{colortbl} 
\usepackage{xcolor} 
\usepackage[skins]{tcolorbox}

\title{Attention-Based Guided Structured Sparsity of Deep Neural Networks}

\author{Amirsina Torfi \\
Virginia Tech\\
\texttt{atorfi@vt.edu} \\
\And
Rouzbeh A.~Shirvani \\
Howard University \\
\texttt{rouzbeh.asgharishir@bison.howard.edu} \\
\And
Sobhan Soleymani \\
West Virginia University \\
\texttt{ssoleyma@mix.wvu.edu} \\
\And
Nasser M. Nasrabadi \\
West Virginia University\\
\texttt{nasser.nasrabadi@mail.wvu.edu} \\
}

\begin{document}

\maketitle

\begin{abstract}
Network pruning is aimed at imposing sparsity in a neural network architecture by increasing the portion of zero-valued weights for reducing its size regarding energy-efficiency consideration and increasing evaluation speed. In most of the conducted research efforts, the sparsity is enforced for network pruning without any attention to the internal network characteristics such as unbalanced outputs of the neurons or more specifically the distribution of the weights and outputs of the neurons. That may cause severe accuracy drop due to uncontrolled sparsity. In this work, we propose an attention mechanism that simultaneously controls the sparsity intensity and supervised network pruning by keeping important information bottlenecks of the network to be active. On CIFAR-10, the proposed method outperforms the best baseline method by $6\%$ and reduced the accuracy drop by $2.6\times$ at the same level of sparsity.

\end{abstract}

\section{Introduction}

%Non-zero parameters are considered to be effective ones. However, one may view this from another perspective by considering the effective parameters to be sufficiently larger ones compared to other parameters.
The main incentive behind model pruning is to impose sparsity by considerably reducing the number of effective parameters in a deep neural network while the accuracy drop is negligible~\citep{han2015deep,denil2013predicting}. Different effective methods such as utilizing group lasso for learning sparse structure~\citet{yuan2006model}, constrain the structure scale~\citet{liu2015sparse}, and regularizing multiple DNN structures known as Structured Sparsity Learning~(SSL)~\citep{wen2016learning} have been implemented for network pruning. 

Unfortunately, there is a lack of addressing two issues for most of the conducted research efforts. First, pruning over-parameterized models with negligible accuracy drop, does not provide rigorous empirical proof for the effectiveness of the model since one can claim manually reducing the network size can generate relatively similar results~\citep{zhu2017prune}. Second, imposing uncontrolled sparsity on under-parameterized baseline models may cause severe accuracy drop. Even if the network is over-parameterized, then imposing two much sparsity may cause the aforementioned issues. 

%So there is a necessity for proposing a guided-pruning mechanism to tackle the problem of accuracy drop caused by over-sparsifying the model.

In this work, we propose a controller mechanism for network pruning with the goal of (1) model compression for having few active parameters by enforcing group sparsity, (2) preventing the accuracy drop by controlling the sparsity of the network using an additional loss function by forcing a portion of the output neurons to stay alive in each layer of the network, and (3) capability of being incorporated for any layer type. Our source code is available online\footnote{\url{https://github.com/astorfi/attention-guided-sparsity}}.

%The proposed gradual adaptive method can be incorporated into the training process for any layer type and demonstrate its effectiveness for model structure learning. 

\section{Attention mechanism for group sparse regularization}

%For having a having a better demonstration of our proposed mechanism, we focus to prune convolutional layers. At first, a mechanism is proposed for preventing accuracy drop and then it will be specified how it will be leveraged to activate more important channel\footnote{A channels in a convolutional layer is correspondent to a neuron for its equivalent fully-connected layer.}.

\begin{figure}[h]
\begin{center}
%\framebox[4.0in]{$\;$}
\includegraphics[scale=0.27]{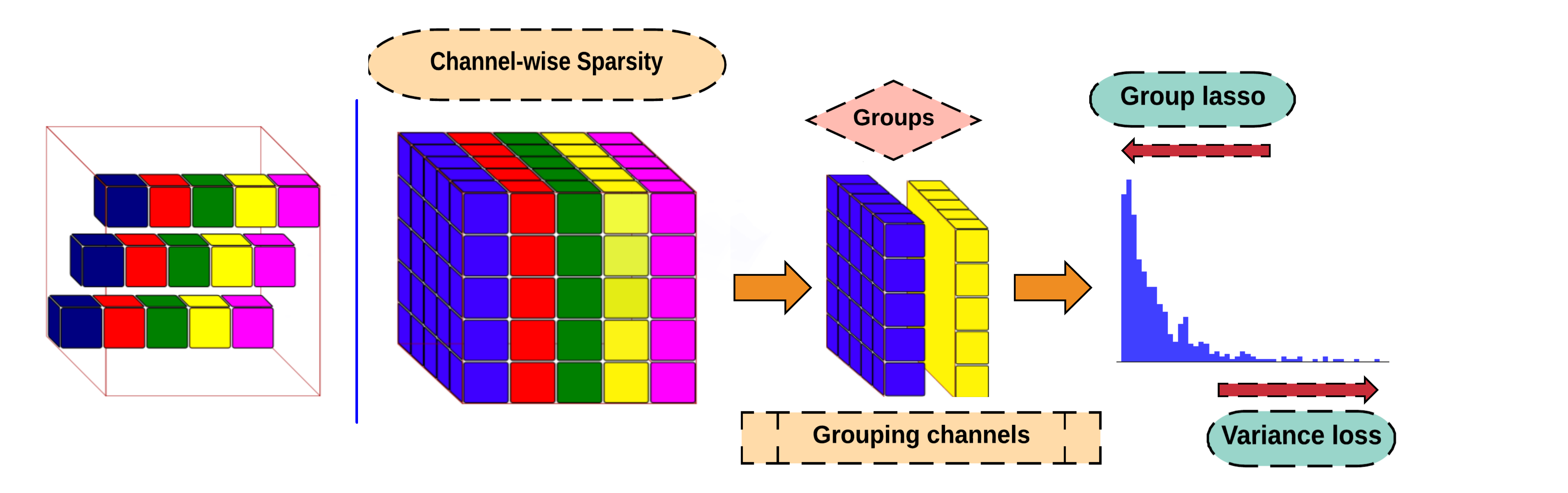}
%\fbox{\rule[-.5cm]{0cm}{4cm} \rule[-.5cm]{4cm}{0cm}}
\end{center}
\caption{Channel-wise grouping and enforcing sparsity in addition to variance loss for each channel.}
\label{fig:varianceloss}
\end{figure}

The weights in a convolutional layer form a tensor as $W\in R^{C,[Width,Heigth],F}$ in which $C$ is the number of input-channel, $[Width,Heigth]$ is the spatial size of the kernel, and $F$ is the number of output filters~(channels). In our proposed method, the objective is the minimization of the following loss function:

\begin{equation}
L(W) = L_{Softmax}(W) + \lambda_{r}.\ell_{2}(W) + \frac{1}{\sqrt{|G(W^{l})|}}\{\lambda_{gs}.\sum_{l=1}^{N}L_{gs}(G(W^{l})) + \lambda_{gv}.\sum_{l=1}^{N}L_{gv}^{-1}(G(W^{l}))\}
\end{equation}

In the above equation, superscript $l$ indicates the layer index\footnote{In the range of [1:N] in case of having N layers.}, $L_{Softmax}(W)$ is the Softmax loss, $\ell_{2}(W)$ is the $\ell_{2}$-regularization loss, and $L_{gs}$ and $L_{gv}$ are the group sparsity and group variance losses respectively. The value of $|G(W^{l})|$ is essentially the number of channels for $l_{th}$ layer and $\lambda$ parameters are the hyper-parameter coefficients for the associated losses. The group sparsity regularization on a set of weights $w$ which are split into M groups can be shown as follows:

\begin{equation}
L_{gs} = \sum_{j=1}^{M}\sqrt{\sum_{i=1}^{|w^{(j)}|}(w_{i}^{(j)})^{2}}
\end{equation}

in which $w^{(j)}$ is the $j_{th}$ group of partial weights in $w$ and $|w^{(j)}|$ is the number of weights in the associated group. Group sparsity has been employed due to its ability for deactivating neurons\footnote{Channels in convolutional layer} by forcing the weights in a group to become zero\footnote{This effectively deactivate the neuron by canceling its output}~\citep{yuan2006model,meier2008group}. The loss function objective leverages group variance loss in addition to group sparsity loss to \textit{force the distribution of the grouped weights to be skewed}. In another word, this attention mechanism, \textit{simply emphasize on a high variance with a concentration around zero}. This will supervise the sparsity mechanism to deliberately keep a portion of grouped weights to be much larger than the majority of the groups in order to simultaneously sparse the architecture and prevent the accuracy drop. Intuitively, this operation forces a portion of channels to be active for transferring sufficient information through the channels in the whole architecture~(information bottlenecks). The visualization of this reasoning is demonstrated in Fig.~\ref{fig:varianceloss}. So basically, in a convolutional layer, each group is all set of weights which forms an output channel.~Equivalently, in a fully-connected layer, a group is the set of outgoing~(ingoing) weights from a neuron. The group-variance is defined as below:

\begin{equation}\label{eq:group-variance}
L_{gv} = \frac{1}{M}\sum_{j=1}^{M}\left (  \sqrt{\sum_{i=1}^{|w^{(j)}|}(w_{i}^{(j)})^{2}} - \frac{1}{M}\sum_{k=1}^{M}\sqrt{\sum_{i=1}^{|w^{(k)}|}(w_{i}^{(k)})^{2}}\right )^{2}
\end{equation}

In case of enforcing sparsity of output channels of convolutional layers, $W_{F_{j}}^{(i)}$ is the $j_{th}$ output channel of the $i_{th}$ layer, then the $L_{gs} = \sum_{j=1}^{N_{filters}}\sqrt{\sum (W_{F_{j}}^{(i)})^{2}}$ and so the formulation of $L_{gv}$ becomes straightforward. We call our method \textit{Guided Structured Sparsity~(GSS)} as it can be considered as an extension to SSL~\citet{wen2016learning} by having an attention mechanism using variational loss that is utilized for supervision of sparsity enforcement operation.

% Moreover, we generalized our method to be used for simultaneously pruning fully-connected and convolutional layers.

\section{Experimental results}

We evaluated our proposed method on two databases: MNIST~\citet{lecun2010mnist}, CIFAR-10~\citet{krizhevsky2009learning}.In all our experiments we enforce the sparsity on both fc-layers~(Using group sparsity for neurons inputs) and convolutional layers~(Using channel-wise structured sparsity for eliminating unimportant filters). In the experiment on MNIST dataset, an architecture similar to LeNet  \citet{lecun1998gradient} has been utilized as the baseline for investigation of our proposed method with no data augmentation. For experiments on CIFAR-10 dataset, we use the \textit{ConvNet} provided by TensorFlow~\citet{tensorflow2015-whitepaper}. The utilized baseline model contains two convolutional layers with \textit{Local Response Normalization~(LRN)}~\citet{krizhevsky2012imagenet} followed by two fully connected layers\footnote{Further details: \url{https://www.tensorflow.org/tutorials/deep_cnn}}.

% for having a considerable reduction in the number of weights and an increase in the speed, respectively.

%~Training starts from randomly initialized weights and sparsity is measured by the portion of weights forced to be zero. It can be equivalent to the portion of channels and neurons which have been turned off in the convolutional and fully-connected sections of the architecture. 

%~By applying our proposed method, for the convolutional layers we enforce network pruning with solely filter-wise sparsity as we did not find it to be more effective to constrain the architecture with both filter-wise and channel-wise sparsity.

%achieves the error rate of \%15.5 using data augmentation. It
%For data augmentation, random cropping, random horizontal flipping, and image whitening have been performed.

\begin{table}[h]
\caption[Table caption text]{Error rate for Different Methods on MNIST and CIFAR-10 at the same level of sparsity~(90\%).}
\label{table:comprison}
\begin{center}
\addtolength{\tabcolsep}{-3pt}
\begin{tabular}{cccc}
\toprule 
Method &   & Error(\%) & \\
\hline
\midrule
\rowcolor{black!0}  & \tcbox[enhanced,size=fbox,fontupper=\small\bfseries,
    sharp corners]{MNIST} &  & \tcbox[enhanced,size=fbox,fontupper=\small\bfseries,
    sharp corners]{CIFAR-10} \\
\midrule
\rowcolor{black!5} Baseline~[\textit{no sparsity}]  & 0.93 &  & 15.51 \\ 
\rowcolor{black!10} $\ell_{1}-regularization$  & 3.16 &   & 24.84 \\
\rowcolor{black!15} Network Pruning~\citep{han2015learning} & 2.67 &   & 23.12 \\
\rowcolor{black!15} Sparsely-connected networks~\citep{ardakani2016sparsely} & 1.91 &   & 17.12 \\
\rowcolor{black!20} SSL~\citep{wen2016learning} & 1.43 &   & 18.71 \\
\hline
\rowcolor{black!3} \textbf{Guided Structured Sparsity [ours]} & \textbf{1.21} &   & \textbf{16.13} \\

\bottomrule
\end{tabular}
\end{center}

\end{table}

Table.~\ref{table:comprison} demonstrates the comparison results. It demonstrates that our method achieves less error rate compared to other methods.

%. In the
%table, the baseline is enforcing no sparsity and the others methods are compared to GSS method at same strength of sparsity regularization. The results

\begin{figure}[h]
\begin{center}
%\framebox[4.0in]{$\;$}
\includegraphics[scale=0.22]{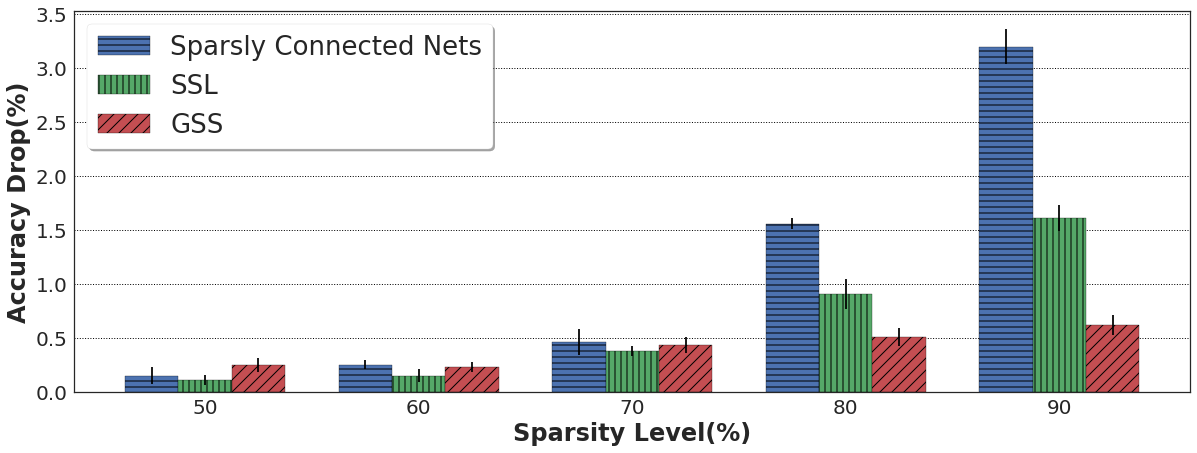}
%\fbox{\rule[-.5cm]{0cm}{4cm} \rule[-.5cm]{4cm}{0cm}}
\end{center}
\caption{The results for experiments on CIFAR-10 dataset at different sparsity levels. 
%The superiority of our proposed method can be observed in higher level of sparsity stengths.
}
\label{fig:sparsitycomparison}
\end{figure}

Fig.~\ref{fig:sparsitycomparison} depicts a comparison at different levels of sparsity. As it can be observed from the figure, our method demonstrates its superiority in higher levels of sparsity.

%for that enforcing a portion of non-sparse connections in a supervised manner becomes critical.

\section{Conclusion}

We have demonstrated that by utilization of the attention mechanism for sparsity supervision, a reduction of $2.6\times$ in accuracy drop has been obtained. Group sparse regularization has been employed on both convolutional and fully-connected layers for simultaneously imposing sparsity and demonstration of the adaptability of the proposed mechanism to both layer types. We anticipate greater superiority of our proposed method compared to the others by utilizing more complex models and evaluation on larger datasets. Besides, it is expected to show advancements in applications such as multi-modality fusion for which network pruning becomes of great importance due to the large number of weights and difficulties in learning a shared common feature space for all modalities~\citep{ngiam2011multimodal,zhao2015heterogeneous}.

\section{Acknowledgement}

This work is based upon a work supported by the Center for Identification Technology Research~(CITeR) and the National Science Foundation~(NSF) under Grant \#1650474.

\bibliography{iclr2018_workshop}
\bibliographystyle{iclr2018_workshop}

\end{document}